\documentclass[conference]{IEEEtran}
\IEEEoverridecommandlockouts
\usepackage{cite}
\usepackage{amsmath,amssymb,amsfonts}
\usepackage{algorithmic}
\usepackage{graphicx}
\usepackage{textcomp}
\usepackage{xcolor}
\usepackage{multirow}
\usepackage{booktabs}
\usepackage{caption}
\usepackage{hyperref}
\usepackage{enumitem}
\usepackage{pifont}
\usepackage{cite}
\usepackage{amsmath, amssymb, amsthm, bm}
\usepackage{float}

\def\BibTeX{{\rm B\kern-.05em{\sc i\kern-.025em b}\kern-.08em
    T\kern-.1667em\lower.7ex\hbox{E}\kern-.125emX}}

\title{
  Multimodal Representation Learning Techniques for Comprehensive Facial State Analysis}

\begin{document}

\author{
    Kaiwen Zheng$^{1}$,
    Xuri Ge$^{2,*}$\thanks{* Corresponding author.},
    Junchen Fu$^{1}$,
    Jun Peng$^{3}$,
    Joemon M. Jose$^{1}$\\
    {\normalsize
    $^{1}$University of Glasgow, School of Computing Science, Glasgow, United Kingdom}\\
    {\normalsize
    $^{2}$Shandong University, School of Artificial Intelligence, Shandong, China}\\
    {\normalsize
    $^{3}$Peng Cheng Laboratory, Shenzhen, China}\\
    {\normalsize
    \texttt{k.zheng.1@research.gla.ac.uk, xuri.ge@sdu.edu.cn, j.fu.3@research.gla.ac.uk,}}\\
    {\normalsize
    \texttt{pengjun.cn@outlook.com, joemon.jose@glasgow.ac.uk}}
}

\maketitle
\begin{abstract}
Multimodal foundation models have significantly improved feature representation by integrating information from multiple modalities, making them highly suitable for a broader set of applications. However, the exploration of multimodal facial representation for understanding perception has been limited. Understanding and analyzing facial states, such as Action Units (AUs) and emotions, require a comprehensive and robust framework that bridges visual and linguistic modalities. In this paper, we present a comprehensive pipeline for multimodal facial state analysis. First, we compile a new Multimodal Face Dataset (MFA) by generating detailed multilevel language descriptions of face, incorporating Action Unit (AU) and emotion descriptions, by  leveraging GPT-4o. Second, we introduce a novel Multilevel Multimodal Face Foundation model (\textbf{MF$^2$}) tailored for Action Unit (AU) and emotion recognition. Our model incorporates comprehensive visual feature modeling at both local and global levels of face image, enhancing its ability to represent detailed facial appearances. This design aligns visual representations with structured AU and emotion descriptions, ensuring effective cross-modal integration. Third, we develop a Decoupled Fine-Tuning Network (DFN) that efficiently adapts MF$^2$ across various tasks and datasets. This approach not only reduces computational overhead but also broadens the applicability of the foundation model to diverse scenarios. Experimentation show superior performance for AU and emotion detection tasks.
\end{abstract}


\section{INTRODUCTION}
\label{sec:intro}
Face representation learning plays an important role in automatic facial state analysis, such as expression recognition \cite{lei2021micro} and medical diagnosis \cite{Jin2022Diagnosis}, and has received extensive attention in recent decades. Its main goal is to extract facial appearance representations for face perception and recognition.
However, face representation learning is very challenging due to the complex and diverse appearance details of facial texture and muscle states. 

Earlier studies \cite{zheng_2022_survey} extracted facial representations from global images using convolutional neural networks (CNNs) to predict facial states such as emotions. For example, Burkert et al. \cite{Fathallah2017Facial} designed a deep CNN for facial expression recognition that uses convolutional layers to capture hierarchical features. While such global representations effectively encode coarse-grained texture and muscle combinations, they often lack the fine-grained localization needed for many downstream tasks. Other works \cite{Zhi2020SV} have focused on facial muscle analysis through Action Unit (AU) recognition, with methods such as \cite{ge2021local,ge2023algrnet} proposing local-global relational networks that accurately locate AU-specific regions via landmark detection. Although both global and local face representations have been successfully applied in tasks like AU recognition \cite{2019yang} and emotion recognition \cite{2018Mehta}, they still do not provide explicit facial feature explanations—for instance, linguistic descriptions—that could further enhance interpretability.

Recently, multimodal joint representation learning has achieved notable success in various applications such as health assessment \cite{Zhou2019MMNG} and driver fatigue detection \cite{SHI2023BSPC}. However, its impact on facial state analysis remains limited due to the complexity of facial appearance features and privacy concerns. On one hand, generating high-quality multimodal face annotations is challenging. Although pre-trained Multimodal Large Language Models (MLLMs) like CoCa \cite{yu2022coca} and Blip \cite{li2022blip} can produce image descriptions in diverse scenarios, no unified approach exists for generating optimal facial state descriptions. Methods such as Exp-BLIP \cite{Yuan_2023_BMVC} and VL-FAU \cite{ge2024towards} use LLMs to generate general face descriptions; however, they either lack sufficiently detailed AU annotations or omit nuanced emotion reasoning. On the other hand, effectively aligning multi-level multimodal face representations—integrating both local and global visual features with corresponding AU and emotion language representations—remains underexplored. For instance, Exp-BLIP \cite{Yuan_2023_BMVC} employs coarse-grained image-text pairs for expression captioning, while VL-FAU \cite{ge2024towards} relies on fixed-form AU descriptions that limit further improvement in visual representation.

In this paper, we address two key challenges in multimodal face representation learning: (i) developing robust, multilevel face annotation methods that provide language-image pairs at various granularities (e.g., detailed AU and emotion context descriptions), and (ii) effectively aligning these multimodal representations to enhance feature extraction.

To this end, we propose a comprehensive pipeline consisting of a novel Multilevel Multimodal Facial Foundation model (MF$^2$) and an efficient Decoupled Fine-Tuning Network (DFN) for downstream tasks. Specifically, we first leverage the pre-trained MLLM GPT-4o \cite{ray_2023_chatgpt} to generate fine-grained AU descriptions and emotion reasoning for face images. Next, the MF$^2$ model integrates local and global visual features with detailed language annotations to yield explicit and comprehensive facial representations, serving as a foundation for tasks such as FAU and emotion recognition. Finally, the DFN enables efficient adaptation of MF$^2$, significantly reducing training overhead while maintaining performance.

The contributions of this paper are as follows:
\begin{itemize}
    \item To enable comprehensive face representation learning, we compile a new multimodal face dataset with high-quality, multilevel language annotations, including descriptions for various AU and emotion reasoning.

    \item We propose a novel Multilevel Multimodal Face Foundation model (MF$^2$) for comprehensive face state analysis, including FAU and emotion recognition. MF$^2$ leverages local and global facial appearance information, aligning it with detailed AU descriptions and reasoning-based emotion annotations.
    
    \item We further provide a fine-tuning method for MF$^2$, referred to as the efficient Decoupled Fine-Tuning Network (DFN), enabling rapid adaptation to new data and enhancing practicality.
\end{itemize}

Extensive experiments on the new multimodal benchmark validate the motivation and effectiveness of our foundation model MF$^2$ and fine-tuning strategy DFN, facilitating the future research of face state analysis.

\section{MULTIMODEL FACIAL ANNOTATION}

\begin{figure}[t]
  \centering
  \vspace{-1em}
  \includegraphics[width=1\linewidth]{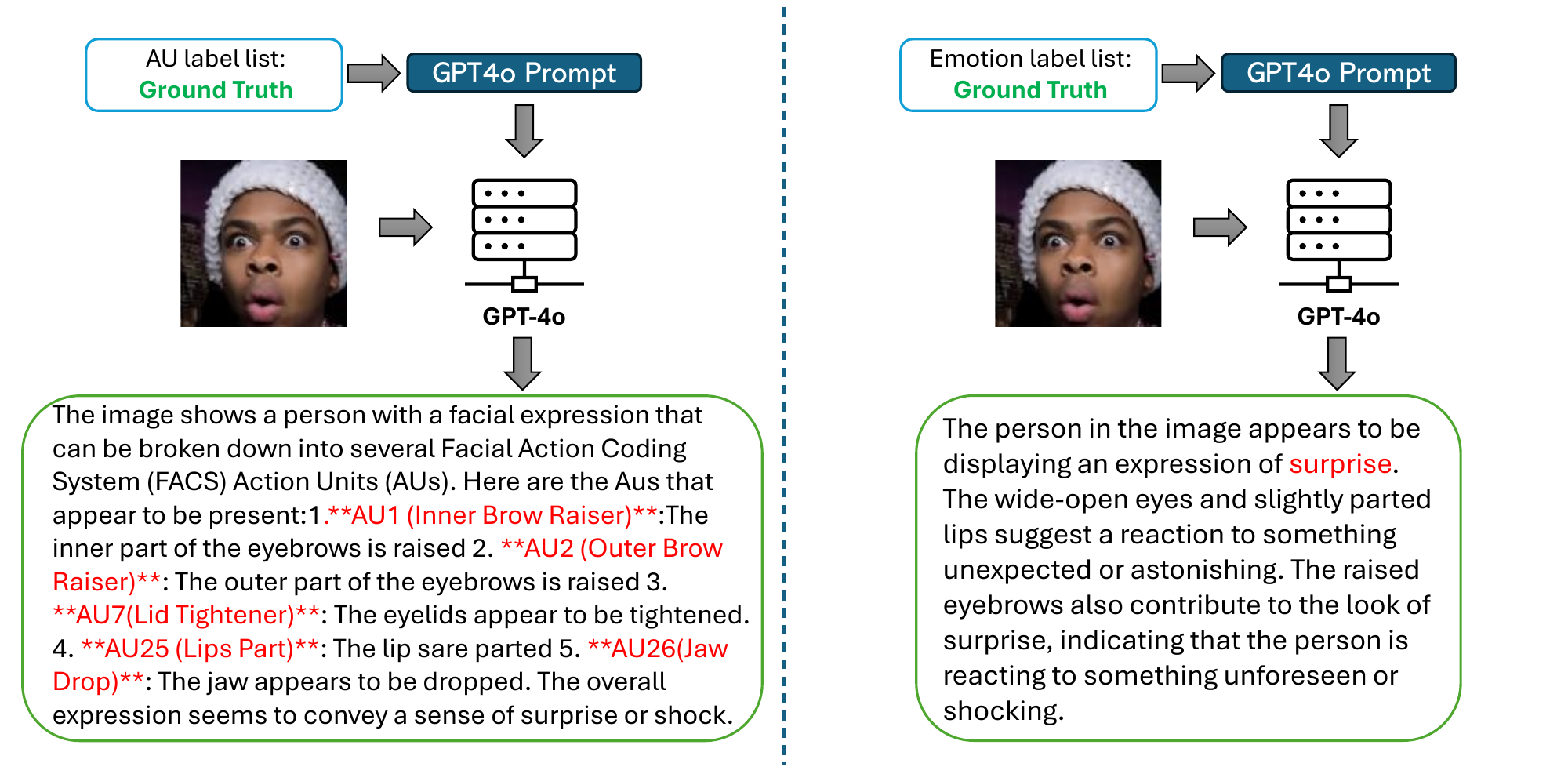}
  \caption{Multimodal facial annotation for detailed AU descriptions and emotion reasoning language based on GPT-4o \cite{ray_2023_chatgpt}. More details are given in supplementary materials.}
  \label{fig:Caption Strategy}
  \vspace{-1em}
\end{figure}

To address the limitations of existing facial datasets, we constructed a new Multimodal Facial dataset (MFA). Figure \ref{fig:Caption Strategy} illustrates the specific steps we followed to reconfigure the dataset, utilizing ground truth labels (emotion and AU annotation) and carefully designed prompts to generate reasonable, high-quality, multilevel facial language descriptions through GPT-4o \cite{ray_2023_chatgpt}. In this section, we introduce the collection process of the dataset, the prompt strategies, and an overview of the MFA dataset.
\subsection{Dataset Construction}

Creating a new dataset from scratch was deemed impractical due to the significant costs and complexities involved. Instead, we opted to use an existing dataset as our foundation. To identify a suitable dataset, we defined two key criteria:
\begin{itemize} 
    \item The dataset must include both Action Unit (AU) and Emotion annotations.
    \item Each image should have an individual emotion label. 
\end{itemize}

After a comprehensive comparison of available datasets, as summarized in Table \ref{tab:dataset_overview2}, we found that only the Aff-Wild2 dataset satisfied these requirements \cite{kollias_2023_affwild2}. Consequently, we selected Aff-Wild2 as the base for our work.

\begin{table}[t]
\centering
\caption{Dataset overview: Comparison between existing  datasets and our constructed dataset.}
\small
\renewcommand{\arraystretch}{0.8} 
\setlength{\tabcolsep}{6pt}
\begin{tabular}{l|c|c|c|c}
\textbf{Name}         & \textbf{AU} &\textbf{Emotion} &\textbf{Requirements} &\textbf{Caption}\\ \hline 
AffectNet \cite{Mollahosseini_2019_AffectNet}      & \ding{55}    & \checkmark   & \checkmark & \ding{55} \\ 
RAF-DB  \cite{li_2019_RAFDB}       & \ding{55}    & \checkmark   & \checkmark  & \ding{55}   \\ 
DFEW  \cite{jiang_2020_dfew}         & \ding{55}    & \checkmark   & \checkmark  & \ding{55}    \\ 
DISFA  \cite{Mavadati_2013_DISFA}        & \ding{55}    & \checkmark   & \checkmark & \ding{55}     \\ 
FERV39K  \cite{wang_2022_ferv39k}      & \ding{55}    & \checkmark   & \checkmark & \ding{55}     \\ 
SFEW \cite{Dhall_2011_SFEW}          & \ding{55}    & \checkmark   & \checkmark  & \ding{55}    \\ 
AFEW  \cite{Kossaifi_2017_AFEW}         & \ding{55}    & \checkmark   & \checkmark & \ding{55}   \\ 
GFT  \cite{Girard_2017_GFT}          & \checkmark    & \checkmark  & \ding{55}& \ding{55}  \\ 
RAF-AU  \cite{Yan_2020_RAFAU}       & \checkmark    & \checkmark  & \ding{55} & \ding{55}  \\ 
CK+  \cite{Lucey_2010_CK}          & \checkmark    & \checkmark  & \ding{55} & \ding{55}\\ 
EmotioNet \cite{Benitez_2016_EmotioNet}     & \checkmark    & \checkmark  & \ding{55} & \ding{55}\\ 
CASME-II   \cite{Qu_2016_CASME2}    & \checkmark    & \checkmark  & \ding{55} & \ding{55} \\ 
BP4D   \cite{ZHANG_2014_BP4D}        & \checkmark    & \checkmark  & \ding{55} & \ding{55}\\ 
AffWild2  \cite{kollias_2023_affwild2}     & \checkmark    & \checkmark  & \checkmark  & \ding{55} \\
\hline 
\textbf{MFA (Ours)} & \checkmark    & \checkmark  & \checkmark & \checkmark  \\ 
\end{tabular}
\label{tab:dataset_overview2}
\end{table}

\noindent \textbf{Data Filtering:} 
To construct a balanced dataset, we began by filtering the Aff-Wild2 dataset to include only images with both Action Units (AUs) and Emotion annotations. This filtering step also ensured emotion class balance across the dataset. Following this process, the refined dataset was split into training and validation sets. Given the video-based nature of the Aff-Wild2 dataset, we maintained a balance in both the number of videos and individual images when dividing the data into these subsets.

\noindent \textbf{GPT-4o Prompt Strategy:} Our objective is to linguistically annotate each image for Action Units (AUs) and emotion, leveraging the existing annotations effectively. Textual descriptions are incorporated to bridge the gap between annotations and model understanding, guiding emotion and AU detection models by highlighting the nuanced differences in these units. This approach helps the models capture subtle variations, improving overall classification accuracy.
To ensure optimal output quality, we experimented with various generation methods and prompt designs. Ultimately, GPT-4o was selected for its nuanced understanding and adaptability. Our structured prompt framework, designed for generating high-quality captions, consists of three key components: task setup, output formatting, and signal specification. This structured approach enables the model to fully comprehend the task, ensuring consistent and detailed outputs across diverse captioning scenarios. Supplementary materials show more prompt design details.
\begin{figure*}[t]
  \centering
  \includegraphics[width=0.9\linewidth]{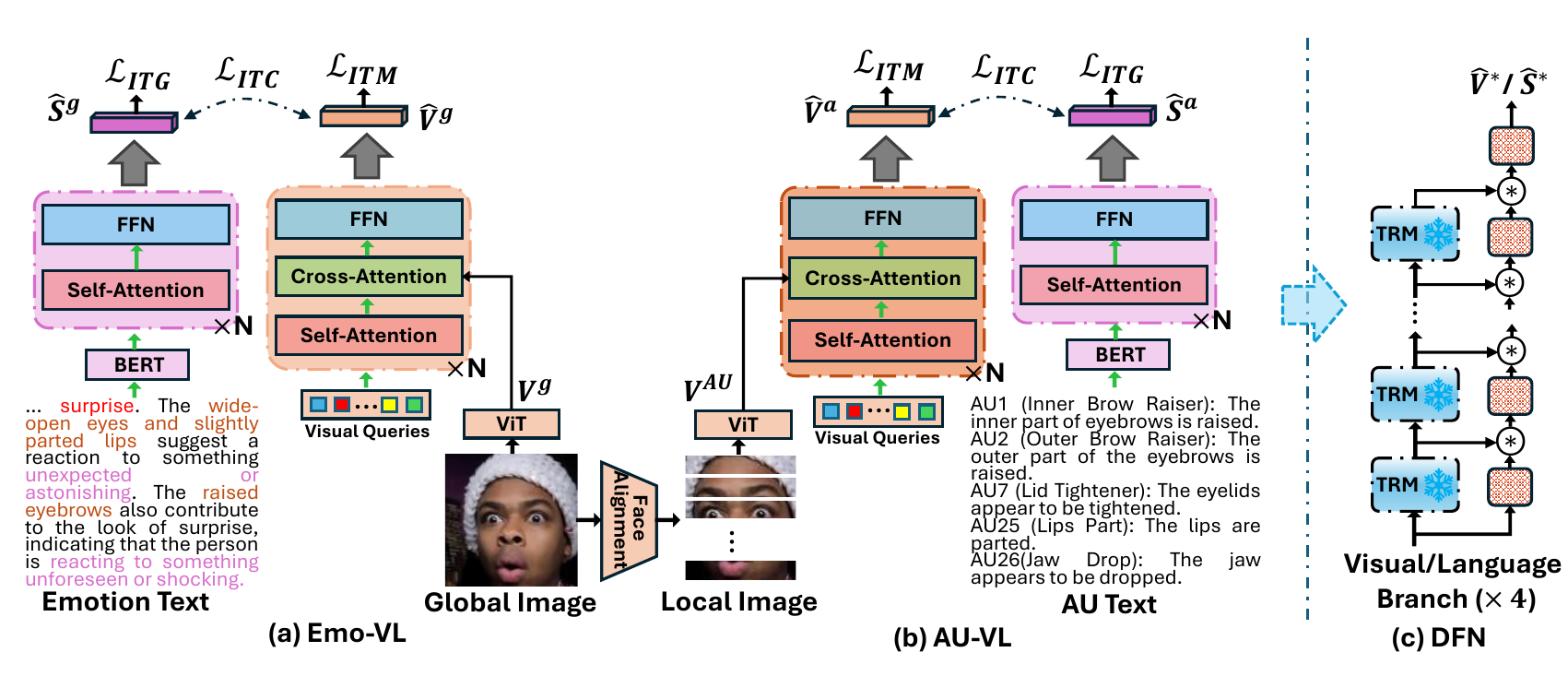}
  \caption{\small Framework: (a) Emo-VL combines global image features with sentiment text; (b) AU-VL integrates local image features with AU text; (c) DFN uses modality-specific side adapters for efficient fine-tuning.}
  \label{fig:Framework}

\end{figure*}

\subsection{Dataset Consolidation and Summarisation}
The dataset comprising a total of 34,696 images extracted from 151 videos. These images have been split into a training set (31,320 images) involving 134 videos and a validation set (3,376 images) involving 17 videos. The data set includes a balanced number of images in eight emotional categories: Neutral, Anger, Disgust, Fear, Happiness, Sadness, Surprise, and Other. Each category has a nearly equal representation in both the training and validation sets to avoid class imbalance, ensuring that the model can generalize well to different emotions. The data set supports three types of caption generation tasks: Emotion Caption, AU Caption, and Key AU Caption. See supplementary material for details of each type of caption.

The extracted and reconstructed dataset, referred to as MFA, is a balanced dataset designed to provide a rich training ground for both AU and emotion classification, ensuring that models trained on it are exposed to diverse facial expressions and action units. 

\section{METHODOLOGY}
We introduce a novel approach for training and fine-tuning a comprehensive multimodal face representation foundation model, illustrated in Figure \ref{fig:Framework}. Our proposed Multilevel Multimodal Face Foundation model (MF$^2$) is designed for diverse facial state analyses, such as FAU and emotion recognition. MF$^2$ leverages newly constructed AU and emotion language descriptions, in MFA, to align with both local and global facial representations, enabling the generation of face representations enriched with detailed features and contextual information.
Furthermore, we propose a new Decoupled Fine-Tuning Network (DFN) for efficiently fine-tuning tasks after training {MF}$^2$. 

\subsection{Multilevel Multimodal Face Foundation Model -- \textbf{{MF}$^2$} }

\textbf{Overview}. 
{MF}$^2$ consists of two main Q-former-based visual-language branches, i.e. a global-level visual encoder with reasoning-based emotion language alignment (Emo-VL) and a local-level visual encoder with fine-grained AU language alignment (AU-VL). The former uses global contexts and situational reasoning in emotional language to assist and improve the ability and discriminability of global face visual representation. The latter further uses each AU language description to accurately improve the visual representation of each muscle area, and improves the fine-grained face representation. During training, we leverage linguistic descriptions to guide the model in identifying situational cues.

\textbf{Emo-VL.}
Following BLIP-2 \cite{Li_2023_blip2}, we model the global face visual representation and emotion description language representation in a unified Q-former, as shown in Figure \ref{fig:Framework} (a).
Emo-VL employs a pre-trained ViT model \cite{dosovitskiy2021vit} to extract the global face feature $V^g$ and then it is input into a Q-former-based multimodal alignment module to align with the emotion language representation $S^e$ from a pre-trained BERT \cite{devlin2019bertl} for the final recognition tasks. 
Specifically, the Q-former-based global alignment module contains a visual encoder and a language encoder. The visual encoder consists of N transformer-based blocks, each containing a Self-Attention layer ($SA$), a Cross-Attention layer ($CA$), and a Feedforward Network ($FFN$). The language encoder also consists of N blocks, where each contains a self-attention layer and an FFN. Due to the characteristics of Q-former \cite{Li_2023_blip2}, an additional cross-attention layer with the learned queries ($Q^g$) is contained in the visual encoder. Similar with BLIP-2 \cite{Li_2023_blip2}, we utilize the Image-Text Contrastive Learning loss ($\mathcal{L}_{\text{ITC}}$), Image-grounded Text Generation loss ($\mathcal{L}_{\text{ITG}}$) and Image-Text Matching loss ($\mathcal{L}_{\text{ITM}} $) to optimise the visual-language alignment and recognition of face states, such as FAU activation state and emotion category, by corresponding task classifiers.
The overall working flow of Emo-VL is formulated as:
\begin{align}
       \hat{V}^g &= FFN(CA(SA(V^g), Q^g)), \\
       \hat{S}^g &= FFN(SA(S^e))
\end{align}
The object functions are followed as: 
\begin{align}
\small
\mathcal{L}_{\text{ITC}}
= -\frac{1}{2M}\sum_{i=1}^M \biggl[ &\log \frac{\exp\bigl(\text{sim}(\mathbf{\hat{v}}^{g}_{i}, \mathbf{\hat{s}}^{e}_{i})/\tau\bigr)}{\sum_{j=1}^M \exp\bigl(\text{sim}(\mathbf{\hat{v}}^{g}_{i}, \mathbf{\hat{s}}^{e}_{j})/\tau\bigr)} \notag \\
&+ \log \frac{\exp\bigl(\text{sim}(\mathbf{\hat{s}}^{e}_{i}, \mathbf{\hat{v}}^{g}_{i})/\tau\bigr)}{\sum_{j=1}^M \exp\bigl(\text{sim}(\mathbf{\hat{s}}^{e}_{i}, \mathbf{\hat{v}}^{g}_{j})/\tau\bigr)} \biggr]
\normalsize
\end{align}

\begin{align}
\small
\mathcal{L}_{\text{ITM}} 
&= -\sum_{i=1}^M \bigl[ y_i \log p(y=1|\mathbf{\hat{v}}^{g}_{i},\mathbf{\hat{s}}^{e}_{i}) \notag \\
&\quad + (1-y_i)\log p(y=0|\mathbf{\hat{v}}^{g}_{i},\mathbf{\hat{s}}^{e}_{i}) \bigr]
\normalsize
\end{align}
\begin{align}
\small
\mathcal{L}_{\text{ITG}} 
&= -\sum_{i \in \text{mask}} \log p(w_i^*|w_{\text{mask}\setminus i}, \mathbf{\hat{V}}^{g})
\normalsize
\end{align}
where $M$ is the size of image-text pairs.  $w_i$ is the target word to predict in text generation or masked language modeling tasks. $\tau$ is a temperature parameter.   

\textbf{AU-VL.}
Emo-VL improves the ability to represent global faces by aligning the global face feature with the emotion language, which explicitly contains global emotion reasoning information. To further compensate for the lack of fine-grained face representation, we propose local face representation enhancement based on the positioning accuracy advantage of Action Units (AUs), as shown in Figure \ref{fig:Framework} (b), named AU-VL. Similarly, we use the AU language description to align with the local AU visual representation in a Q-former-based module to improve its multimodal representation capability. The local AU visual representations are extracted based on the detected face landmarks from a pre-trained landmark detector\cite{bulat2017far}.
The structure of the local Q-former-based alignment module is the same as Emo-VL.
Specifically, to extract the precise AU features in a face image, we use a pre-trained landmark detector \cite{bulat2017far} to localise the AU positions and extract the corresponding representations $V^{AU}$ = $\{V^a_1, ..., V^a_n\}$ from ViT-based visual features. All AU captions are embedded by the BERT \cite{devlin2019bertl} as $S^{AU}$=$\{S^a_1, ..., S^a_n\}$. 
After that, we also employ the Q-former-based AU alignment module to align the local AU visual features and fine-grained AU language features by the same objective functions in Emo-VL. Note that, the visual encoder and language encoder in Q-former alignment are shared for different AUs to save parameters. 
Finally, we obtain the local AU representations $\hat{V}^a$ and their corresponding detailed language representations $\hat{S}^a$. 

During the multilevel visual-language joint learning, we use the cross entropy loss function \cite{mao2023crossentropy} to optimize an AU recognizer and an emotion recognizer respectively for the final facial state analysis.
Thus, we obtain a  face foundation model for FAU recognition and emotion recognition. 

\begin{table*}[t]
\centering

\caption{Quantitative evaluation of AU recognition on the MFA dataset. 
The evaluation metric is F1-score (\%)}
\label{tab:AU_Evaluation}
\setlength{\tabcolsep}{7pt}
\renewcommand{\arraystretch}{1.35}
\begin{tabular}{c|cccccccccccc|c}
\hline
\textbf{Models} & \textbf{AU1} & \textbf{AU2} & \textbf{AU4} & \textbf{AU6} & \textbf{AU7} & \textbf{AU10} & \textbf{AU12} & \textbf{AU15} & \textbf{AU23} & \textbf{AU24} & \textbf{AU25} & \textbf{AU26} & \textbf{Avg.} \\
\hline
Exp-BLIP \cite{Yuan_2023_BMVC} & 40.25 & 12.63 & 63.41 & \underline{53.28} & 69.43 & 71.76 & \textbf{60.18} & 46.85 & 27.60 & 10.27 & 86.43 & 25.61 & 47.31 \\

ME-GraphAU \cite{Luo_2022_IJCAI} & 41.94 & 13.72 & 55.91 & 41.92 & 76.57 & 70.48 & 53.68 & \textbf{61.42} & 20.13 & 03.88 & 85.53 & 30.47 & 46.30 \\

VL-FAU \cite{ge2024towards} & 43.09 & \underline{15.86} & 55.59 & 49.35 & 77.57 & \textbf{73.51} & 54.81 & \underline{60.00} & \underline{29.50} & 03.72 & 84.25 & 31.08 & 48.19 \\

\specialrule{0.1pt}{0pt}{0pt}
MF$^2$ (Pre-Train) & \textbf{50.17} & \textbf{18.75} & \textbf{73.18} & \textbf{54.83} & \underline{76.58} & 70.00 & 52.57 & 48.92 & 29.06 & \underline{11.72} & \underline{88.68} & \textbf{34.80} & \underline{50.77} \\

MF$^2$ (Fine-Tuning) & \underline{44.76} & 15.64 & \underline{66.90} & 50.42 & \textbf{76.70} & \underline{73.17} & \underline{57.80} & 54.51 & \textbf{33.49} & \textbf{43.02} & \textbf{89.26} & \underline{34.55} & \textbf{53.35} \\
\hline
\end{tabular}
\end{table*}
\subsection{Efficient Decoupled Fine-tuning Network -- \textbf{DFN}}
As the foundational backbone of MF$^2$, the Q-former faces two primary limitations: (1) its transformer-based architecture is computationally expensive, and (2) to mitigate this cost, it employs shared Self-Attention and FFN modules for multimodal contrastive learning. While this design may enhance cross-modal interaction, it compromises the unique representation capability of individual modalities. To address these challenges and improve the generalization of the proposed foundation model MF$^2$, we propose a simple yet effective Decoupled Fine-Tuning Network (DFN) for pre-trained MF$^2$  built entirely with lightweight adapters. The detailed framework is shown in Figure \ref{fig:Framework} (c).
Inspired by the advanced Side Adapter paradigm \cite{fu2024efficient,fu2024iisan}, which outperforms traditional adapters and LoRA in efficiency \cite{fu2024exploring,houlsby2019parameter}, DFN decouples the shared modules into distinct side adapter pathways. By incorporating unique modality-specific adjustments through two independent side adapters, DFN effectively mitigates interference between modalities while significantly reducing computational overhead.
Specifically, DFN is parallel to each modality branch in MF$^2$ and performs decoupling fine-tuning. Therefore, there are 4N DFN cells in total, each of which consists of a downsampling and upsampling layer composed of a fully connected layer, and is connected using an activation function. 
When fine-tuning the DFN, we freeze the MF$^2$ backbone and only update the parameters of DFN for the new task, under the optimization of new task objective functions.

\begin{table*}[t]
\centering

\caption{Quantitative evaluation of emotion recognition on the MFA dataset.
The evaluation metric is  accuracy (\%)}
\label{tab:Emotion_Evaluation}
\setlength{\tabcolsep}{10pt}
\renewcommand{\arraystretch}{1.35}
\begin{tabular}{c|cccccccc|c}
\hline
\textbf{Model} & \textbf{Neutral} & \textbf{Anger} & \textbf{Disgust} & \textbf{Fear} & \textbf{Happiness} & \textbf{Sadness} & \textbf{Surprise} & \textbf{Other}  & \textbf{Avg.} \\
\hline
Exp-BLIP \cite{Yuan_2023_BMVC} & 82.17 & \underline{92.74} & \underline{86.58} & \textbf{86.79} & \textbf{90.73} & \underline{88.30} & 79.56 & 50.24 & 82.14  \\

HSEmotion \cite{SAVCHENKO2022} & 80.95 & 85.99 & \textbf{86.82} & \underline{86.73} & 85.31 & 77.84 & 69.05 & \underline{76.30} & 81.12  \\

\specialrule{0.1pt}{0pt}{0pt}

MF$^2$ (Pre-Train) & \textbf{87.70} & \textbf{95.50} & 86.37 & 86.64 & \underline{86.05} & 88.09 & 82.08 & 55.39 & \underline{83.48}  \\

MF$^2$ (Fine-Tuning) & \underline{84.53} & 92.57 & 79.95 & 82.41 & 83.92 & \textbf{89.51} & \underline{87.14} & 75.21 & \textbf{84.40}  \\

\hline
\end{tabular}
\end{table*}

\begin{table*}[t]
\centering

\caption{Ablation analysis of emotion recognition Model.
The evaluation metric is accuracy (\%), TT for training time (min/epoch), IT for inference time (min/epoch), and TP for trainable parameters}
\label{tab:Ablation_Study}
\setlength{\tabcolsep}{5pt}
\renewcommand{\arraystretch}{1.35}
\begin{tabular}{c|cccccccc|c|c|c|c}
\hline
\textbf{Model} & \textbf{Neutral} & \textbf{Anger} & \textbf{Disgust} & \textbf{Fear} & \textbf{Happiness} & \textbf{Sadness} & \textbf{Surprise} & \textbf{Other}  & \textbf{Avg.} & \textbf{TT}& \textbf{IT}& \textbf{TP} \\
\hline
MF$^2$ (Fine-Tuning) & 84.53 & 92.57 & 79.95 & 82.41 & \underline{83.92} & \textbf{89.51} & \textbf{87.14} & \underline{75.21} & \textbf{84.40} & 12.6 & 6.4 & 52.88M\\

\specialrule{0.1pt}{0pt}{0pt}

w/o DFN & 87.70 & \textbf{95.50} & 86.37 & 86.64 & \textbf{86.05} & 88.09 & \underline{82.08} & 55.39 & \underline{83.48} & 62.3 & 5.1 & 373.4M\\

w/o Emo-VL & \underline{88.57} & \underline{94.55} & \underline{86.82} & \underline{86.76} & 84.66 & 84.86 & 50.71 & \textbf{78.44} & 82.42 & 8.4 & 4.5 & 186.7M\\

w/o AU-VL & \textbf{90.52} & 93.39 & \textbf{86.85} & \textbf{87.17} & 71.39 & \underline{88.92} & 71.18 & 61.56 & 81.87 & 4.1 & 2.1 & 186.7M\\

\hline
\end{tabular}
\vspace{-0.2in}
\end{table*}

\section{EXPERIMENTS}
\subsection{Experimental Settings}
 \noindent \textbf{Implemental Details.} All details are shown in supplementary materials.

\noindent \textbf{Evaluation Metrics.} 
The evaluation metrics include the F1 score for facial Action Unit (AU) detection and the classification accuracy for face emotion recognition.

\subsection{Experimental Results}

\noindent \textbf{Compared Methods.} We compare the proposed MF$^2$ and its DFN-based fine-tuning model with three baselines for AU recognition in Table \ref{tab:AU_Evaluation} and two baselines for emotion recognition in Table \ref{tab:Emotion_Evaluation}.  For AU recognition, it contains ME-GraphAU \cite{Luo_2022_IJCAI}, Exp-BLIP \cite{Yuan_2023_BMVC}, VL-FAU \cite{ge2024towards}. For Emotion recognition, HSEomtion \cite{SAVCHENKO2022} and Exp-BLIP \cite{Yuan_2023_BMVC} are compared with our models. More baseline model details are shown in supplementary materials.

\noindent \textbf{Performance of FAU Recognition.} 
Table \ref{tab:AU_Evaluation} highlights the performance of various models on the MFA dataset for FAU recognition. Among the baseline models, VL-FAU achieves the highest average performance with an F1 score of 48.19\%. However, both versions of our proposed MF$^2$ model significantly outperform these baselines. Specifically, the MF$^2$ (Pre-Train) model achieves an average F1 score of 50.77\%, while MF$^2$ (Fine-Tuning) further improves to 53.35\%, representing a substantial margin of +5.16\% over the best-performing baseline (VL-FAU).

\noindent \textbf{Performance of Emotion Recognition.} 
Table \ref{tab:Emotion_Evaluation} presents the performance of various models on the MFA dataset for emotion recognition. 
Our MF$^2$ (Pre-Train) model achieves an average accuracy of 83.48\%, and the MF$^2$ (Fine-Tuning) model further boosts performance to 84.40\%, demonstrating a notable margin of +2.26\% over the best-performing baseline Exp-BLIP\cite{Yuan_2023_BMVC}. These results, combined with the recognition results from the FAU, highlight the comprehensive capabilities of the MF$^2$ model. By utilising the Emo-VL and AU-VL modules, MF$^2$ effectively integrates both global and fine-grained facial features aligned with corresponding diverse AU and emotion language, ensuring superior performance across different tasks. Furthermore, the success of the MF$^2$ (Fine-Tuning) model demonstrates the effectiveness of decoupling the DFN implementation. Overall, this highlights the robustness and adaptability of the model in multimodal facial representation learning.

\subsection{Ablation Study}

\noindent To demonstrate the effectiveness of the proposed modules, we conducted extensive ablation studies. We show how each component influences the overall performance of the MF$^2$ model. Table \ref{tab:Ablation_Study} presents the component ablation study for the MF$^2$ model, including (1) Efficiency of Decoupled Fine-Tuning and (2) Impact of Global and Local Feature Integration.

\noindent \textbf{Efficiency of Decoupled Fine-Tuning (DFN).}  Removing the Decoupled Fine-Tuning Network (DFN) led to a performance drop of 0.92\% and increased training time from 12.6 minutes per epoch to 62.3 minutes, as shown in Table \ref{tab:Ablation_Study}. Moreover, the number of trainable parameters rose drastically from 52.88 million with DFN to 373.42 million without it. These findings underscore DFN's critical role in reducing computational overhead and optimizing parameter efficiency while maintaining high performance.

\noindent \textbf{Impact of Global (Emo-VL) and Local (AU-VL) Feature Integration.} Furthermore, removing the AU-VL module resulted in a significant performance drop (-2.53\%), compared to a smaller drop (-1.98\%) when the Emo-VL module was removed, as shown in Table \ref{tab:Ablation_Study}. Additionally, training time decreased to 4.1 minutes per epoch without AU-VL and to 4.5 minutes without Emo-VL, highlighting a trade-off between computational efficiency and model effectiveness. These results demonstrate that AU-VL plays a pivotal role in capturing fine-grained, muscle-specific features, while Emo-VL enhances global contextual understanding. Together, these modules ensure a balanced and comprehensive facial representation.

The ablation studies confirm the effectiveness of the MF$^2$ model's design, highlighting the critical role of each component in achieving state-of-the-art performance while ensuring computational and parameter efficiency.

\section{CONCLUSION}
This paper presented a novel multimodal facial representation learning pipeline, integrating image and text modalities to enhance AU and emotion recognition. We compiled the {MFA} dataset with high-quality detailed AU  and emotional description linguistically. The proposed foundation model MF$^2$ effectively combines global (Emo-VL) and local (AU-VL) visual-language representations with emotion and AU language alignment learning, ensuring comprehensive and detailed facial feature enhancement. Additionally, our Decoupled Fine-Tuning Network (DFN) enables efficient task-specific fine-tuning, reducing computational cost and achieving superior performance. 
Experimental results validated the effectiveness of our multimodal MF$^2$ model and its efficient fine-tuning strategy (DFN), outperforming state-of-the-art methods while demonstrating a reduction in training time. 
Future work will focus on exploring advanced multimodal representations and improving relational reasoning in face analysis.

\bibliographystyle{IEEEbib}
\bibliography{icme2025}

\begin{thebibliography}{10}

\bibitem{lei2021micro}
Ling Lei, Tong Chen, Shigang Li, and Jianfeng Li,
\newblock ``Micro-expression recognition based on facial graph representation learning and facial action unit fusion,''
\newblock in {\em CVPR}, 2021, pp. 1571--1580.

\bibitem{Jin2022Diagnosis}
Bo~Jin, Leandro Cruz, and Nuno Gonçalves,
\newblock ``Deep facial diagnosis: Deep transfer learning from face recognition to facial diagnosis,''
\newblock {\em IEEE Access}, vol. 8, pp. 123649--123661, 2020.

\bibitem{zheng_2022_survey}
Yinglin Zheng, Hao Yang, Ting Zhang, Jianmin Bao, Dongdong Chen, Yangyu Huang, Lu~Yuan, Dong Chen, Ming Zeng, and Fang Wen,
\newblock ``General facial representation learning in a visual-linguistic manner,'' 2022.

\bibitem{Fathallah2017Facial}
Abir Fathallah, Lotfi Abdi, and Ali Douik,
\newblock ``Facial expression recognition via deep learning,''
\newblock in {\em AICCSA}, 2017, pp. 745--750.

\bibitem{Zhi2020SV}
Ruicong Zhi, Mengyi Liu, and Dezheng Zhang,
\newblock ``A comprehensive survey on automatic facial action unit analysis,''
\newblock {\em Vis. Comput.}, vol. 36, no. 5, pp. 1067–1093, May 2020.

\bibitem{ge2021local}
Xuri Ge, Pengcheng Wan, Hu~Han, Joemon~M Jose, Zhilong Ji, Zhongqin Wu, and Xiao Liu,
\newblock ``Local global relational network for facial action units recognition,''
\newblock in {\em FG}. IEEE, 2021, pp. 01--08.

\bibitem{ge2023algrnet}
Xuri Ge, Joemon~M Jose, Pengcheng Wang, Arunachalam Iyer, Xiao Liu, and Hu~Han,
\newblock ``Algrnet: Multi-relational adaptive facial action unit modelling for face representation and relevant recognitions,''
\newblock {\em IEEE Transactions on Biometrics, Behavior, and Identity Science}, vol. 5, no. 4, pp. 566--578, 2023.

\bibitem{2019yang}
Jiannan Yang, Fan Zhang, Bike Chen, and Samee~U. Khan,
\newblock ``Facial expression recognition based on facial action unit,''
\newblock in {\em IGSC}, 2019, pp. 1--6.

\bibitem{2018Mehta}
Dhwani Mehta, Mohammad Faridul~Haque Siddiqui, and Ahmad~Y. Javaid,
\newblock ``Facial emotion recognition: A survey and real-world user experiences in mixed reality,''
\newblock {\em Sensors}, vol. 18, no. 2, 2018.

\bibitem{Zhou2019MMNG}
Tao Zhou, Mingxia Liu, Kim-Han Thung, and Dinggang Shen,
\newblock ``Latent representation learning for alzheimer’s disease diagnosis with incomplete multi-modality neuroimaging and genetic data,''
\newblock {\em ITMT}, vol. 38, no. 10, pp. 2411--2422, 2019.

\bibitem{SHI2023BSPC}
Jinxuan Shi and Kun Wang,
\newblock ``Fatigue driving detection method based on time-space-frequency features of multimodal signals,''
\newblock {\em BSPC}, vol. 84, pp. 104744, 2023.

\bibitem{yu2022coca}
Jiahui Yu, Zirui Wang, Vijay Vasudevan, Legg Yeung, Mojtaba Seyedhosseini, and Yonghui Wu,
\newblock ``Coca: Contrastive captioners are image-text foundation models,'' 2022.

\bibitem{li2022blip}
Junnan Li, Dongxu Li, Caiming Xiong, and Steven Hoi,
\newblock ``Blip: Bootstrapping language-image pre-training for unified vision-language understanding and generation,'' 2022.

\bibitem{Yuan_2023_BMVC}
Yujian Yuan, University of~Chinese Academy~of Science, Jiabei Zeng, and Shiguang Shan,
\newblock ``Describe your facial expressions by linking image encoders and large language models,''
\newblock in {\em BMVC 2023}. 2023, BMVA.

\bibitem{ge2024towards}
Xuri Ge, Junchen Fu, Fuhai Chen, Shan An, Nicu Sebe, and Joemon~M Jose,
\newblock ``Towards end-to-end explainable facial action unit recognition via vision-language joint learning,''
\newblock in {\em ACM MM}, 2024, pp. 8189--8198.

\bibitem{ray_2023_chatgpt}
Partha~Pratim Ray,
\newblock ``Chatgpt: a comprehensive review on background, applications, key challenges, bias, ethics, limitations and future scope,''
\newblock {\em Internet of Things and Cyber-Physical Systems}, vol. 3, pp. 121--154, 04 2023.

\bibitem{kollias_2023_affwild2}
Dimitrios Kollias, Panagiotis Tzirakis, Alice Baird, Alan Cowen, and Stefanos Zafeiriou,
\newblock ``Abaw: Valence-arousal estimation, expression recognition, action unit detection \& emotional reaction intensity estimation challenges,'' 2023.

\bibitem{Mollahosseini_2019_AffectNet}
Ali Mollahosseini, Behzad Hasani, and Mohammad~H. Mahoor,
\newblock ``Affectnet: A database for facial expression, valence, and arousal computing in the wild,''
\newblock {\em IEEE TAC}, vol. 10, no. 1, pp. 18–31, Jan. 2019.

\bibitem{li_2019_RAFDB}
Shan Li and Weihong Deng,
\newblock ``Reliable crowdsourcing and deep locality-preserving learning for unconstrained facial expression recognition,''
\newblock {\em IEEE TIP}, vol. 28, no. 1, pp. 356--370, 2019.

\bibitem{jiang_2020_dfew}
Xingxun Jiang, Yuan Zong, Wenming Zheng, Chuangao Tang, Wanchuang Xia, Cheng Lu, and Jiateng Liu,
\newblock ``Dfew: A large-scale database for recognizing dynamic facial expressions in the wild,'' 2020.

\bibitem{Mavadati_2013_DISFA}
S.~Mohammad Mavadati, Mohammad~H. Mahoor, Kevin Bartlett, Philip Trinh, and Jeffrey~F. Cohn,
\newblock ``Disfa: A spontaneous facial action intensity database,''
\newblock {\em IEEE TAC}, vol. 4, no. 2, pp. 151--160, 2013.

\bibitem{wang_2022_ferv39k}
Yan Wang, Yixuan Sun, Yiwen Huang, Zhongying Liu, Shuyong Gao, Wei Zhang, Weifeng Ge, and Wenqiang Zhang,
\newblock ``Ferv39k: A large-scale multi-scene dataset for facial expression recognition in videos,'' 2022.

\bibitem{Dhall_2011_SFEW}
Abhinav Dhall, Roland Goecke, Simon Lucey, and Tom Gedeon,
\newblock ``Static facial expression analysis in tough conditions: Data, evaluation protocol and benchmark,''
\newblock in {\em 2011 IEEE ICCV Workshops}, 2011, pp. 2106--2112.

\bibitem{Kossaifi_2017_AFEW}
Jean Kossaifi, Georgios Tzimiropoulos, Sinisa Todorovic, and Maja Pantic,
\newblock ``Afew-va database for valence and arousal estimation in-the-wild,''
\newblock {\em Image Vision Comput.}, vol. 65, no. C, pp. 23–36, Sept. 2017.

\bibitem{Girard_2017_GFT}
Jeffrey~M Girard, Wen-Sheng Chu, L{'{a}}szl{'{o}}~A Jeni, Jeffrey~F Cohn, Fernando {De La Torre}, and Michael~A Sayette,
\newblock ``{Sayette group formation task (GFT) spontaneous facial expression database},''
\newblock in {\em IEEE FG}, 2017.

\bibitem{Yan_2020_RAFAU}
Wen-Jing Yan, Shan Li, Chengtao Que, Jiquan Pei, and Weihong Deng,
\newblock ``Raf-au database: In-the-wild facial expressions with subjective emotion judgement and objective au annotations,''
\newblock in {\em ACCV}, November 2020.

\bibitem{Lucey_2010_CK}
Patrick Lucey, Jeffrey~F. Cohn, Takeo Kanade, Jason Saragih, Zara Ambadar, and Iain Matthews,
\newblock ``The extended cohn-kanade dataset (ck+): A complete dataset for action unit and emotion-specified expression,''
\newblock in {\em CVPR Workshops}, 2010, pp. 94--101.

\bibitem{Benitez_2016_EmotioNet}
C.~Fabian Benitez-Quiroz, Ramprakash Srinivasan, and Aleix~M. Martinez,
\newblock ``Emotionet: An accurate, real-time algorithm for the automatic annotation of a million facial expressions in the wild,''
\newblock in {\em CVPR}, 2016, pp. 5562--5570.

\bibitem{Qu_2016_CASME2}
Fangbing Qu, Sujing Wang, Wen-Jing Yan, and Xiaolan Fu,
\newblock ``Cas(me)2: A database of spontaneous macro-expressions and micro-expressions,''
\newblock in {\em Interacci{\'o}n}, 2016.

\bibitem{ZHANG_2014_BP4D}
Xing Zhang, Lijun Yin, Jeffrey~F. Cohn, Shaun Canavan, Michael Reale, Andy Horowitz, Peng Liu, and Jeffrey~M. Girard,
\newblock ``Bp4d-spontaneous: a high-resolution spontaneous 3d dynamic facial expression database,''
\newblock {\em IVC}, vol. 32, no. 10, pp. 692--706, 2014,
\newblock Best of Automatic Face and Gesture Recognition 2013.

\bibitem{Li_2023_blip2}
Junnan Li, Dongxu Li, Silvio Savarese, and Steven Hoi,
\newblock ``Blip-2: Bootstrapping language-image pre-training with frozen image encoders and large language models,'' 2023.

\bibitem{dosovitskiy2021vit}
Alexey Dosovitskiy, Lucas Beyer, Alexander Kolesnikov, Dirk Weissenborn, Xiaohua Zhai, Thomas Unterthiner, Mostafa Dehghani, Matthias Minderer, Georg Heigold, Sylvain Gelly, Jakob Uszkoreit, and Neil Houlsby,
\newblock ``An image is worth 16x16 words: Transformers for image recognition at scale,'' 2021.

\bibitem{devlin2019bertl}
Jacob Devlin, Ming-Wei Chang, Kenton Lee, and Kristina Toutanova,
\newblock ``Bert: Pre-training of deep bidirectional transformers for language understanding,'' 2019.

\bibitem{bulat2017far}
Adrian Bulat and Georgios Tzimiropoulos,
\newblock ``How far are we from solving the 2d \& 3d face alignment problem? (and a dataset of 230,000 3d facial landmarks),''
\newblock in {\em ICCV}, 2017.

\bibitem{mao2023crossentropy}
Anqi Mao, Mehryar Mohri, and Yutao Zhong,
\newblock ``Cross-entropy loss functions: Theoretical analysis and applications,'' 2023.

\bibitem{Luo_2022_IJCAI}
Cheng Luo, Siyang Song, Weicheng Xie, Linlin Shen, and Hatice Gunes,
\newblock ``Learning multi-dimensional edge feature-based au relation graph for facial action unit recognition,''
\newblock in {\em IJCAI}. July 2022, IJCAI-2022, p. 1239–1246, IJCAI.

\bibitem{fu2024efficient}
Junchen Fu, Xuri Ge, Xin Xin, Alexandros Karatzoglou, Ioannis Arapakis, Kaiwen Zheng, Yongxin Ni, and Joemon~M Jose,
\newblock ``Efficient and effective adaptation of multimodal foundation models in sequential recommendation,''
\newblock {\em arXiv preprint arXiv:2411.02992}, 2024.

\bibitem{fu2024iisan}
Junchen Fu, Xuri Ge, Xin Xin, Alexandros Karatzoglou, Ioannis Arapakis, Jie Wang, and Joemon~M Jose,
\newblock ``Iisan: Efficiently adapting multimodal representation for sequential recommendation with decoupled peft,''
\newblock in {\em Proceedings of the 47th International ACM SIGIR Conference on Research and Development in Information Retrieval}, 2024, pp. 687--697.

\bibitem{fu2024exploring}
Junchen Fu, Fajie Yuan, Yu~Song, Zheng Yuan, Mingyue Cheng, Shenghui Cheng, Jiaqi Zhang, Jie Wang, and Yunzhu Pan,
\newblock ``Exploring adapter-based transfer learning for recommender systems: Empirical studies and practical insights,''
\newblock in {\em Proceedings of the 17th ACM international conference on web search and data mining}, 2024, pp. 208--217.

\bibitem{houlsby2019parameter}
Neil Houlsby, Andrei Giurgiu, Stanislaw Jastrzebski, Bruna Morrone, Quentin De~Laroussilhe, Andrea Gesmundo, Mona Attariyan, and Sylvain Gelly,
\newblock ``Parameter-efficient transfer learning for nlp,''
\newblock in {\em ICML}. PMLR, 2019, pp. 2790--2799.

\bibitem{SAVCHENKO2022}
Andrey~V. Savchenko,
\newblock ``Hsemotion: High-speed emotion recognition library,''
\newblock {\em Software Impacts}, vol. 14, pp. 100433, 2022.

\bibitem{Ge_2024_VL}
Xuri Ge, Junchen Fu, Fuhai Chen, Shan An, Nicu Sebe, and Joemon~M. Jose,
\newblock ``Towards end-to-end explainable facial action unit recognition via vision-language joint learning,''
\newblock in {\em MM}. Oct. 2024, MM ’24, p. 8189–8198, ACM.

\end{thebibliography}

\clearpage
\appendix
\section{Details of Multimodal Facial Annotation}
\subsection{GPT-4o Prompt Design}

We designed a three-stage GPT-4o prompt (Initial Setup, Output Formpt and Output Signal) to generate the three high-quality descriptive captions we needed: the AU caption, the Emo caption and the Key AU caption. Below, we discuss the rationale and considerations behind the prompt structures used.\\   

\textit{\textbf{Note}: The prompt examples used in the introductory architecture section are all \textbf{emotion prompt examples}.}

\begin{figure}[h]
  \centering
  \includegraphics[width=1\linewidth]{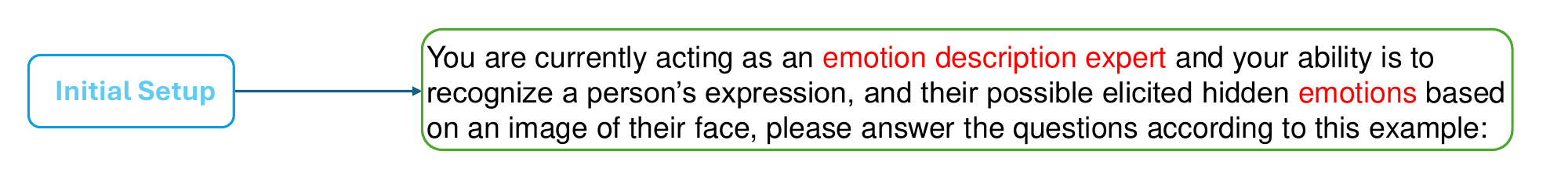}
  \caption{Emotion Initial Setup Prompt}
  \label{fig:Emotion Initial Setup Prompt}
\end{figure}

\textbf{Initial Setup.} In this step Figure \ref{fig:Emotion Initial Setup Prompt}, the Prompt model is assigned a specific role relevant to the task. For example, the model can be instructed to take on the role of an "emotion description expert" or an "action unit recognition expert." This helps ChatGPT better understand the task's context, clarify the desired goal, and focus on a particular task, such as recognizing Action Units in emoticons. By doing so, the model reduces ambiguity and applies relevant knowledge more accurately, enhancing the response's relevance and the quality of the generated results. This step ensures that the model performs optimally when addressing specific problems, thereby effectively improving the accuracy and consistency of the generated content.

\begin{figure}[h]
  \centering
  \includegraphics[width=1\linewidth]{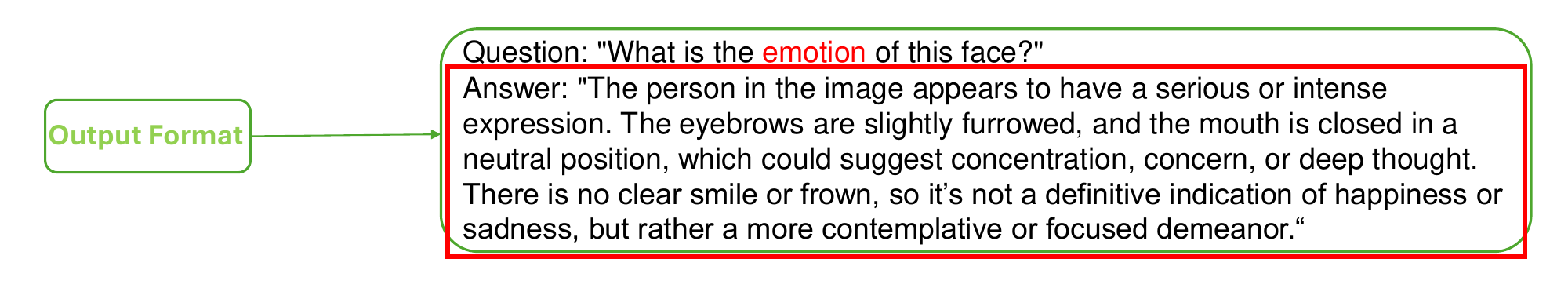}
  \caption{Emotion Output Format Prompt}
  \label{fig:Emotion Output Format Prompt}
\end{figure}

\textbf{Output Format.} In this step Figure \ref{fig:Emotion Output Format Prompt}, we provide the model with an example question-answer format that serves as a guide for structuring its responses. This example helps the model understand the desired level of detail, tone, and format, ensuring standardized outputs across different inputs. By referencing the example, the model learns to include all necessary components in its responses, such as specific facial features, their emotional implications, and the relationships between facial action units. This consistency is especially crucial for complex tasks like emotion and AU classification, where responses must be informative, contextually relevant, and coherent. The example acts as a template, helping the model generate responses that are accurate, well-organized, and easy to interpret. Additionally, it sets a standard for depth and clarity, ensuring that the model consistently delivers context-aware, detailed, and relevant outputs.

\begin{figure}[h]
  \centering
  \includegraphics[width=1\linewidth]{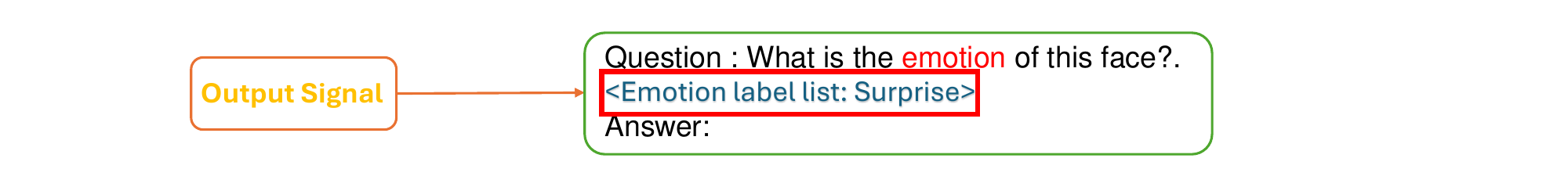}
  \caption{Emotion Output Signal Prompt}
  \label{fig:Emotion Output Signal Prompt}
\end{figure}

\textbf{Output Signal.} In this step Figure \ref{fig:Emotion Output Signal Prompt}, we provide the model with an example question-answer format that serves as a guide for structuring its responses. This example helps the model understand the desired level of detail, tone, and format, ensuring standardized outputs across different inputs. By referencing the example, the model learns to include all necessary components in its responses, such as specific facial features, their emotional implications, and the relationships between facial action units. This consistency is especially crucial for complex tasks like emotion and AU classification, where responses must be informative, contextually relevant, and coherent. The example acts as a template, helping the model generate responses that are accurate, well-organized, and easy to interpret. Additionally, it sets a standard for depth and clarity, ensuring that the model consistently delivers context-aware, detailed, and relevant outputs.

\textbf{Summary.} We employ a novel prompt-based method using GPT-4o \cite{ray_2023_chatgpt} to generate detailed captions for both emotion and action unit (AU) analysis, offering deeper insights into facial expressions and their emotional implications. For emotion captioning, the model, guided by a prompt that positions it as an "emotion description expert," interprets subtle facial cues such as eyebrow or lip movements to produce rich, context-aware descriptions beyond simple emotion labels. For AU captioning, the model acts as an "AU description expert," breaking down facial expressions into specific AUs (e.g., AU4 for Brow Lowerer, AU24 for Lip Pressor) with detailed explanations of their contributions to overall expressions. Furthermore, the key AU caption approach focuses on identifying the most influential AUs for a given emotion, highlighting their decisive roles in conveying emotional states. This integrated approach provides a comprehensive understanding of how facial muscle movements define emotions, offering precise interpretations of complex expressions where multiple AUs interact.

\subsection{GPT-4o Prompt Example}
\begin{figure*}[h]
  \centering
  \includegraphics[width=0.8\linewidth]{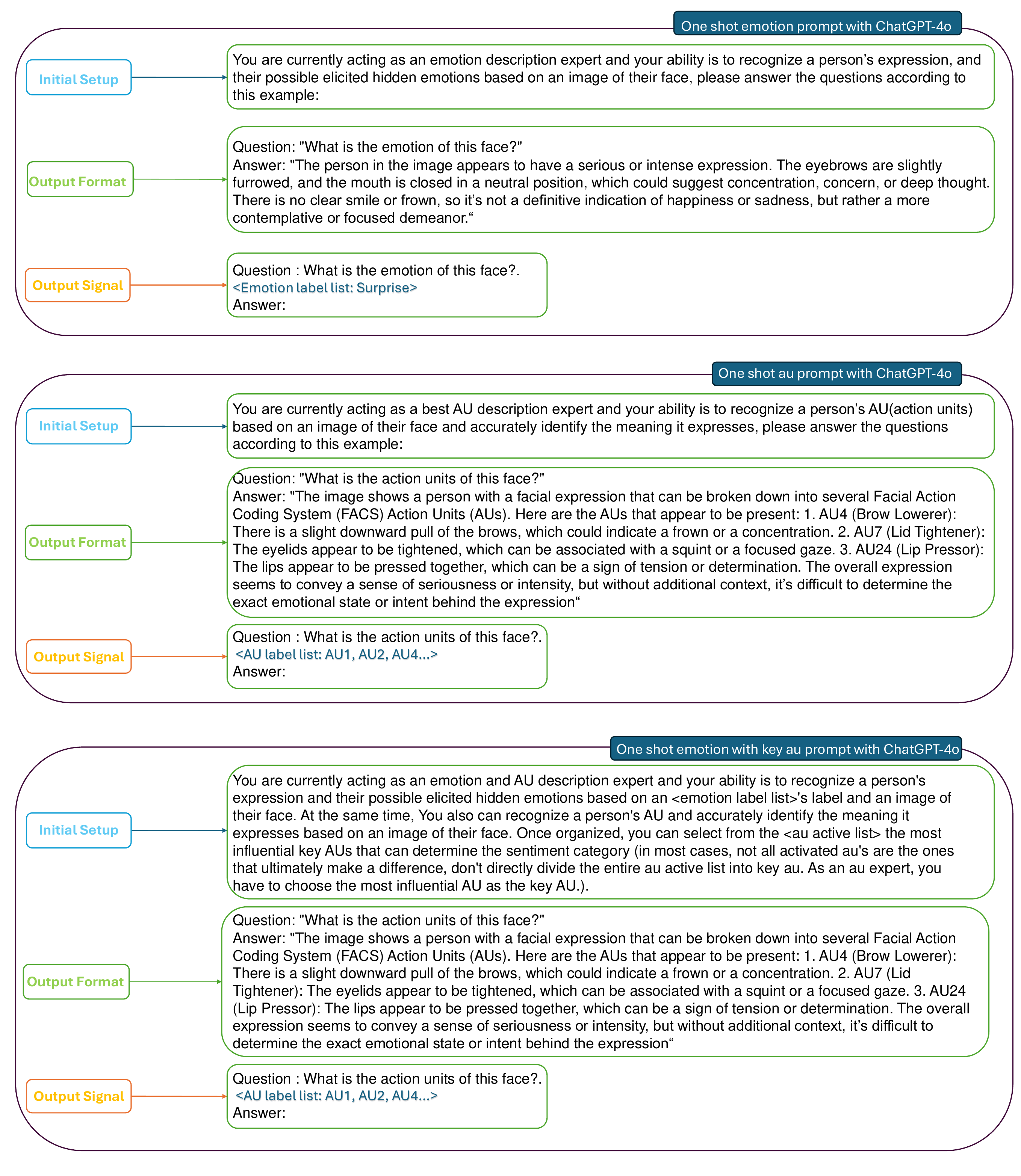}
  \caption{Three Types of Prompt Design Details (AU prompt, emotion prompt and key AU prompt). When you finally type the prompt we will personalize the output format according to the format you want to get, this format is not fixed, it just depends on what information you need to get. }
  \label{fig:Three Types of Prompt Design Details}
\end{figure*}

Due to space limitations in the main text, we cannot present complete examples of the three prompt types and their generated captions (AU caption, emotion caption, and key AU caption). To clarify the differences among these three prompts, we provide basic examples of each in Figure \ref{fig:Three Types of Prompt Design Details}.

As shown in the figure, the differences between AU captions and emotion captions are minimal. The key distinction lies in the initial role setting (emotion expert or AU expert), which ensures GPT focuses on the required domain knowledge while mitigating the influence of unrelated factors. Another difference is the Output Format, which controls the content of the required response and indirectly guides GPT’s reasoning process. In contrast, the key AU caption differs significantly from the other two. It emphasizes the interaction between AUs and emotions and incorporates more detailed prompt settings to achieve this.

In summary, this design ensures GPT-generated captions are accurate, contextually rich, and tailored to support downstream tasks.

\begin{figure*}[h]
  \centering
  \includegraphics[width=0.8\linewidth]{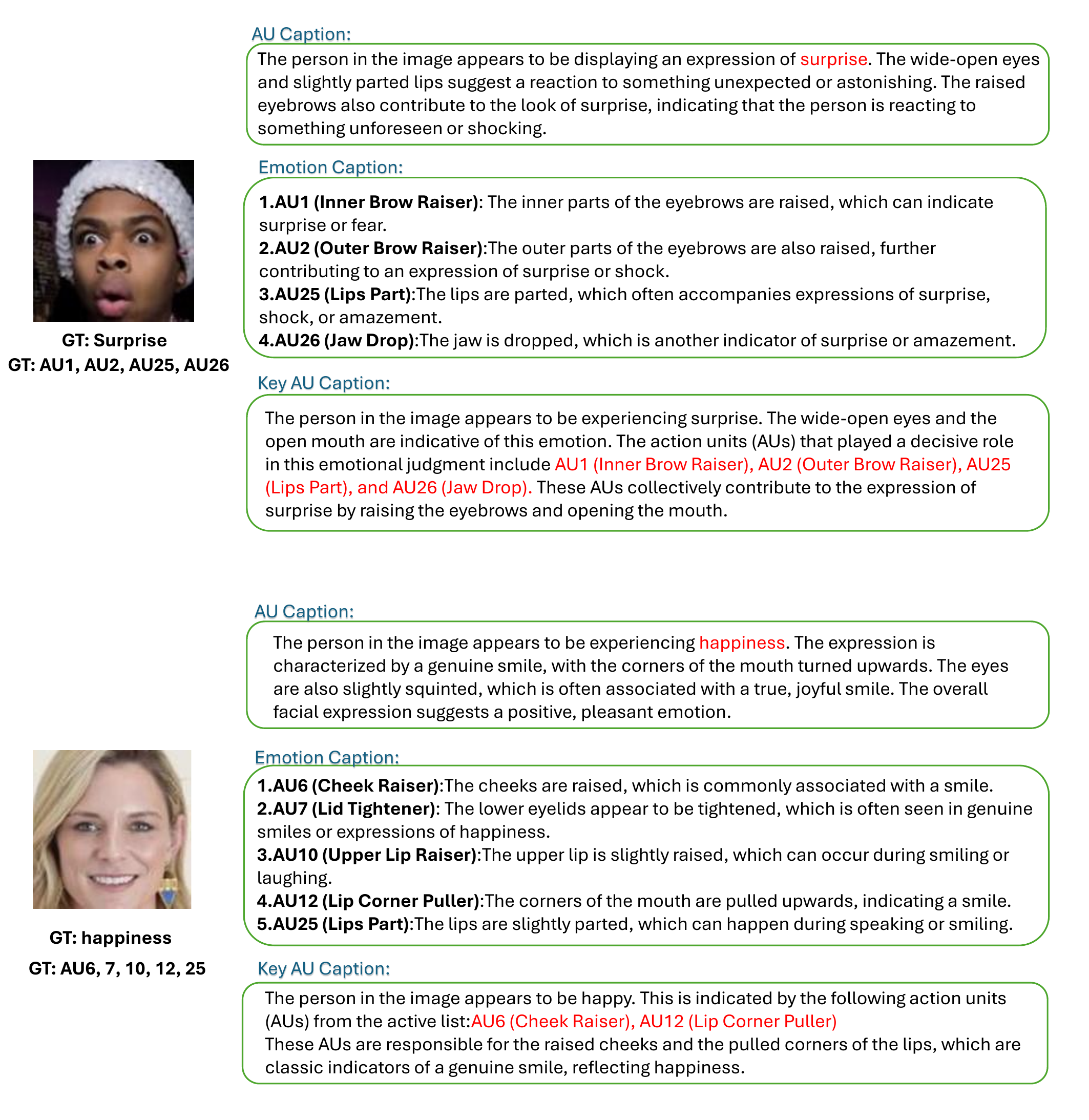}
  \caption{Three Types of Caption Example (AU caption, emotion caption and key AU caption).All captions are entered into the GPT using a combination of the designed prompt and the corresponding ground true, and it is worth noting that the prompt is accompanied by the corresponding image, which allows the GPT to generate a personalized caption for the image.}
  \label{fig:Three Types of Caption Example}
\end{figure*}

\subsection{Example of Caption}
From the two sets of examples in Figure \ref{fig:Three Types of Caption Example} we can clearly see the characteristics of the three different captions. Each type of caption serves a distinct purpose:

\begin{itemize}
    \item \textbf{Emotion Caption}: Provides an overview of the emotional state expressed by the face, utilizing the full spectrum of emotions present in the dataset.
    
    \item \textbf{AU Captiont}: Describes the specific facial action units, breaking down the muscle movements involved in the expression.
    
    \item \textbf{Key AU Caption}: Highlights the most influential action units that determine the emotional state, based on the ground truth emotion and AU labels. This novel caption type helps identify the critical facial movements responsible for conveying specific emotions.
\end{itemize}

Compared to conventional single captions manually generated solely based on Ground Truth Labels, our approach uses refined prompts, images, and Ground Truth Labels as inputs to generate captions through ChatGPT-4o \cite{ray_2023_chatgpt}. This method produces more descriptive captions that incorporate the intrinsic information of the images, resulting in captions that are more accurate, unique, and diverse.

To this end, we designed a key AU prompt that provides a unique approach to generating captions with large language models. When generating key AU captions, only the image, prompt, and the Ground Truth Labels for AU and Emotion are provided, without including any information about the key AU itself. The carefully crafted prompt ensures that ChatGPT-4o \cite{ray_2023_chatgpt} fully analyzes the image, going beyond simplistic descriptions of individual AUs and emotions.

\section{More Experiments and Detailed Hyper-parameter Settings}
\subsection{Transcition Experiments}
Finally, we pre-train AU on the MF$^2$ model and then fine-tune emotion on the MF$^2$ (Fine-tuning) model after pre-training, and we name this final model MF$^2$ (Intern-VL).We tested the performance of this model on emotion and against the baseline model, and according to the results in table \ref{tab:Transition}, we can find that our model is 1.16\% better than Exp-BLIP \cite{Yuan_2023_BMVC}.

\begin{table}[hb]
\centering
\captionsetup{justification=centering, labelsep=period}
\caption{Transition for AU Pre-train to Emotion Fine-tuning}
\label{tab:Transition}
\setlength{\tabcolsep}{10pt}
\renewcommand{\arraystretch}{1.5}
\begin{tabular}{c|c}
\hline
\textbf{Model} & \textbf{Avg} \\
\hline
Exp-BLIP\cite{Yuan_2023_BMVC} & \underline{78.91} \\

\specialrule{0.1pt}{0pt}{0pt}

MF$^2$ (Intern-VL) & \textbf{80.07}  \\

\hline
\end{tabular}
\end{table}

\subsection{Experimental Parameters}
All experiments were conducted on an RTX A6000 GPU. Additional detais on the hyperparameter settings are providd in Table \ref{tab:hyperparams}

\begin{table}[H]
\centering
\caption{Experimental Parameters}
\renewcommand{\arraystretch}{1.2}
\label{tab:hyperparams}
\begin{tabular}{@{}lccc@{}}
\toprule
\multicolumn{4}{c}{\textbf{Common Parameters}} \\ \midrule
\textbf{Parameter} & \multicolumn{3}{c}{\textbf{Value}} \\ \midrule
Training epoch & \multicolumn{3}{c}{30} \\
Optimizer & \multicolumn{3}{c}{AdamW \cite{loshchilov2019adamW}} \\
Weight decay & \multicolumn{3}{c}{0.05} \\
Linear warm-up & \multicolumn{3}{c}{2000 steps} \\
Learning rate & \multicolumn{3}{c}{$1 \times 10^{-4}$} \\
Image size & \multicolumn{3}{c}{224*224} \\
Batch size & \multicolumn{3}{c}{56} \\
\midrule
\multicolumn{4}{c}{\textbf{Multilevel Multimodal Face Foundation Model ($MF^2$)}} \\ \midrule
\textbf{Parameter} & \textbf{AU-VL} & \textbf{Emo-VL} & \\ \midrule
Temp & 0.07 * torch.ones([]) & 0.07 * torch.ones([]) & \\
Caption max length & 169 & 61 & \\
\midrule
\multicolumn{4}{c}{\textbf{Decoupled Fine-tuning Network  (DFN)}} \\ \midrule
\textbf{Parameter} & \textbf{Image} & \textbf{Text} & \\ \midrule
CLS tokens & Last layer of ViT & Last layer of Bert & \\
Number of Adapter Layers & 7 & 7 & \\
Activation Function & ReLU/Sigmoid & ReLU/Sigmoid & \\
Input Dimension & 768 & 768 & \\
Output Dimension & 768 & 768 & \\
Gate Scaling Factor & 0.1 & 0.1 & \\
\bottomrule
\end{tabular}
\end{table}

\subsection{Baseline Model Details}
We compared the results of multiple baseline models on the MFA dataset at both the AU and Emotion levels. Below is a detailed introduction to the baseline models.

\begin{itemize}
    \item\textbf{Exp-BLIP} \cite{Yuan_2023_BMVC} employs a multimodal transformer architecture based on BLIP-2 to integrate image and text modalities. It processes AU and Emotion representations independently, limiting its ability to fully capture their interplay.
    
    \item\textbf{ME-GraphAU} \cite{Luo_2022_IJCAI} utilizes graph neural networks to model relationships between facial regions for AU recognition, effectively enhancing the detection of Action Units through structured interconnections.
    
    \item\textbf{VL-FAU} \cite{Ge_2024_VL} incorporates visual-linguistic representations to improve AU recognition tasks. It focuses on aligning visual features with linguistic cues to achieve state-of-the-art performance.
    
    \item\textbf{HSEmotion} \cite{SAVCHENKO2022} focuses on emotion recognition by classifying emotional states. It achieves competitive results but does not explicitly address the integration of AUs for comprehensive facial analysis.
\end{itemize}

\subsection{Differences in Inputs Between Training and Validation}

During training, the model employs both images and textual descriptions (e.g., emotion and AU captions) to cultivate a richer visual-semantic understanding. Objectives like image-text matching and image-text contrast reinforce multimodal alignment, enabling the model to capture subtle facial expressions and nuanced features more effectively.

In contrast, the validation phase uses images alone for three primary reasons:

\begin{itemize}
    \item\textbf{Realistic Deployment Scenarios} Textual information may be unavailable in practice. Restricting validation to images ensures performance metrics reflect actual application conditions.
    
    \item\textbf{Generalization and Robustness} Evaluating the model without text verifies that it can perform effectively under conditions not explicitly supported during training, confirming its adaptability.

    \item\textbf{Fair and Independent Assessment} Excluding previously seen textual descriptions prevents artificially inflated performance, resulting in a more authentic gauge of the model’s true capabilities.
\end{itemize}

\vspace{12pt}

\end{document}